\documentclass{INTERSPEECH2023}
\usepackage{multirow,makecell,subcaption}
\usepackage{microtype}
\usepackage{relsize}

\interspeechcameraready

\usepackage{stfloats,hyperref}

\newcommand{\mypara}[1]{\vspace{0.5em} \noindent{\textbf{#1}} \hspace{0.25em}}
\newcommand{\testclean}{{\footnotesize \texttt{test-clean}}}
\newcommand{\devclean}{{\footnotesize \texttt{dev-clean}}}
\newcommand{\trainclean}{{\footnotesize \texttt{train-clean-100}}}

\title{Self-supervised Predictive Coding Models Encode Speaker and Phonetic Information in Orthogonal Subspaces
\thanks{This work was supported by the UKRI Centre for Doctoral Training in Natural Language Processing, funded by the UKRI (grant EP/S022481/1) and the University of Edinburgh, School of Informatics and School of Philosophy, Psychology \& Language Sciences.}
}
\name{Oli Danyi Liu, Hao Tang,  Sharon Goldwater}
\address{
  School of Informatics, University of Edinburgh, United Kingdom}
\email{oli.liu@ed.ac.uk, hao.tang@ed.ac.uk, sgwater@inf.ed.ac.uk}

\begin{document}

\maketitle
 
\begin{abstract}
Self-supervised speech representations are known to encode both speaker and phonetic information, but how they are distributed
in the high-dimensional space remains largely unexplored.
We hypothesize that they are encoded in orthogonal
subspaces, a property that lends itself to simple disentanglement. 
Applying principal component analysis to representations of two predictive coding models, we identify
two subspaces that capture speaker and phonetic variances, and confirm that they are nearly orthogonal.
Based on this property, we propose a new speaker normalization method 
which collapses the subspace that encodes speaker information, without requiring transcriptions.
Probing experiments show that our method effectively eliminates speaker information 
and outperforms a previous baseline in phone discrimination tasks.
Moreover,
the approach generalizes and can be used to remove information of unseen speakers.

\end{abstract}


\section{Introduction}
Self-supervised learning (SSL) models of speech are trained on large quantities of unlabeled data to learn useful representations of speech audio. 
Empirically, SSL pre-training has been shown to improve performance on downstream tasks while reducing reliance on annotated data \cite{Zhang2023GoogleUS, polyak21_interspeech}.
Researchers have also started to evaluate SSL models as computational models for phonetic acquisition and speech perception in humans \cite{blandon2020analysis, lavechin2022statistical}. 

It is natural to ask what makes these learned representations so effective. 
So far, most work in the area has done so by analyzing {\em what} type of information is encoded, finding that SSL models encode information ranging from acoustic \cite{wells22_interspeech} and contextual \cite{sanabria_analyzing_2022} cues, to gender \cite{deseyssel22_interspeech} and speaker identity \cite{niekerk21_interspeech}. Some researchers have also analyzed {\em where} in these models (at which layer) such information is encoded \cite{pasad}.
In this paper, we ask a question that has received little attention so far, namely {\em how} different types of information are distributed across the dimensions of the representation space.
Analysis targeted at this question can potentially improve the interpretability of the learned representations, inspire new methods for disentangling linguistic and non-linguistic information, and shed light on properties of the representation space that can be compared against neural representations or behaviour data of humans to evaluate the scientific value of the model.
    We know of no previous work analyzing this question in detail, although some studies have shown through two-dimensional visualizations that representations may cluster according to their corresponding language, gender, or phone class \cite{deseyssel22_interspeech}; word-level context \cite{sanabria_analyzing_2022}; or speaker identity \cite{niekerk21_interspeech}.

In this work, we explicitly investigate how speaker and phonetic information are distributed
in the representation space learned by SSL models. We hypothesize that a good representation (one that is efficient and works well for predicting speech) should implicitly disentangle these two sources of information, since they vary independently in the processes that generate the speech signal. If so, then the two types of information would be represented in orthogonal subspaces of the representation space, and it would be possible to perform speaker normalization by identifying
the speaker subspace and collapsing it.

We test our hypothesis in experiments with two SSL models (APC \cite{chung19_interspeech} and CPC \cite{oord_representation_2019}) trained on English LibriSpeech. 
We use principal component analysis (PCA) to identify the subspaces that capture most of the variance for speakers and for phones, and confirm that these subspaces are nearly orthogonal. 

We then test whether the speaker subspace generalizes to unseen speakers by using it to perform speaker normalization on a test set: we project the representations in the test set onto the subspace orthogonal to the speaker subspace learned from a training set.
We use probing classifiers and an ABX phone discrimination task to show that with the resulting representations, (a) little speaker information remains; and (b) within- and across-speaker phone discrimination improves, outperforming a baseline of utterance-level standardization \cite{niekerk21_interspeech}. Our results suggest that these two SSL models implicitly disentangle speaker and phone information into orthogonal subspaces. 

\section{Overview of the approach}

We demonstrate the orthogonal speaker and phone subspaces in the SSL representations using a simple approach, where we first aggregate the representations by speaker and phone, then perform PCA on the resulting matrices\footnote{Performing PCA on the raw representations is far more compute-intensive and did not give the clear patterns we see in Figure \ref{fig: similarity}; the aggregation step seems necessary to overcome noise.} and compare the principal directions.
We then show that collapsing the speaker subspace is an effective speaker normalization technique, for both seen and unseen speakers. These steps are explained in more detail below.

\mypara{Aggregating representations by speaker and phone} Suppose that the dataset for analysis contains
a set of speakers $S$ and has time-aligned phone transcriptions (which are used for analysis only, and are not needed for speaker normalization).
We denote $Z_s$ to be the set of frame-level representation vectors (each of dimension $D$) from an SSL model of a speaker $s \in S$.
Similarly, we define $Z_p$ to be the set of hidden vectors labeled as phone $p \in P$, where $P$ is a phone set; we define $Z_{s, p}$ to be the set of hidden vectors labeled as phone $p \in P$ and of a speaker $s \in S$.

For our analysis, we aggregate the representations in three ways: (1) by speaker (2) by phone, and (3) by each combination of speaker and phone.
This results in three matrices: 
The first, $M_\text{spk}$, is a $|S| \times D$ matrix of speakers, where the $s$-th row $M_\text{spk}[s]$ is
the average of all frames of a speaker $s \in S$, i.e., $M_{\text{spk}}[s] = \text{avg}(Z_s)$. 
Analogously, $M_\text{phn}$ is a $|P| \times D$ matrix of phones, where the $p$-th row $M_\text{phn}[p]$
is the average of all frames labeled as $p \in P$, i.e., $M_\text{phn}[p] = \text{avg}(Z_p)$.
Finally, $M_{\text{joint}}$ is a $|S||P| \times D$ matrix, where each row corresponds to $\text{avg}(Z_{s, p})$ for a particular speaker $s$ and a particular phone $p$.

\mypara{Identifying speaker and phone dimensions} Applying PCA to $M_\text{spk}$ gives us $|S|$ principal components that describe the directions that account for the largest variance among the speakers.
A \textit{speaker subspace} is the subspace spanned by the eigenvectors corresponding to the top eigenvalues.
Similarly, PCA on $M_\text{phn}$ and $M_{\text{joint}}$ gives us
a phone subspace and a joint speaker-phone subspace, respectively. 

In section \ref{sec: analysis}, we analyze the relationships between these matrices by measuring the similarity between their principal directions.
We also look at the projection of $M_{\text{joint}}$ on each of its principal
directions to explore what specific information is encoded in each dimension.

\mypara{Collapsing the speaker subspace} If speaker and phonetic information are represented in orthogonal
subspaces, projecting frames to the subspace orthogonal to the speaker subspace would remove speaker information without affecting how well phonetic information can be extracted.
Specifically, for a hidden vector $z$ and a principal direction $v$, the vector $z' = z - (z^\top v)v$ is orthogonal to $v$.
This step can be performed on multiple principal directions, and we refer to the projection as collapsing the speaker space when $v$ is a principal component of $M_\text{spk}$.
Before doing this, we need to choose the number of principal components to use, which 
can be done based on the cumulative variance explained by the principal components for the training set, or by tuning development set performance (e.g., on phone discrimination).
To test whether a speaker subspace generalizes across different sets of speakers,
we use a speaker subspace learned from a different set of speakers for collapsing.

\section{Experimental setup}
The two models used in this work are the autoregressive predictive coding (APC) model \cite{chung19_interspeech} and the contrastive predictive coding (CPC) model \cite{oord_representation_2019}.
Both CPC and APC are trained to predict one or several future frame(s) given past context in the same utterance, with the main difference being that CPC is optimized with a contrastive learning objective, \textit{i.e.} to distinguish the true future frame from a set of negative samples, whereas APC is directly optimized to minimize the distance between the predicted and the target frames.
These models, which learn representations by trying to predict unobserved continuous frames in the future, stand in contrast to some other SSL models such as HuBERT \cite{Hsu2021HuBERTSS} and wav2vec 2.0 \cite{Baevski2020wav2vec2A}, which perform masked prediction of quantized units given context on both sides, and which we intend to explore in future.\footnote{We follow recent work in speech technology and machine learning in using the term {\em predictive coding} to refer to error-driven learning based on forward prediction, contrasting with masked prediction or other objectives. In the more general sense used in information theory \cite{elias1955predictive, yang2022autoregressive} and computational neuroscience \cite{huang_predictive_2011, Rao1999PredictiveCI, Friston2005ATO}, masked prediction can also be viewed as a type of predictive coding.
}

We use two CPC models of different sizes and one APC model. 
Both CPC-big and CPC-small are taken from the Zero Speech 2021 baseline\footnote{ \url{https://github.com/zerospeech/zerospeech2021_baseline}}.
We use our own implementation of APC, following \cite{yang2022autoregressive}.
The CPC and APC models differ in several ways.
CPC has a 5-layer convolution that extracts 10-ms frames from wave samples for further processing, while APC operates on 10-ms Log Mel features.
We extract the representations from the LSTM component of the models (also called context network in CPC).
The prediction horizon, as parameterized by $K$, is 12 for CPC and 3 for APC.
The specific hyperparameters are chosen based on prior work \cite{nguyen_zero_2020,chung19_interspeech} and are listed in Table \ref{tab: model}.

\begin{table}
    \centering
    \caption{Summary of model specifications. Models are trained on LibriLight (LL) or LibriSpeech (LS) train-clean.}
    \resizebox{\columnwidth}{!}{\begin{tabular}{lccccc}
    \toprule
        \multirow{2}{*}{Model} & LSTM & Extracted & Hidden & \multirow{2}{*}{$K$} & Training \\
        & layers & layer & units &  & data \\
        \midrule
        CPC-big & 4 & 2 & 512 & 12 & LL 6k hrs \\
        CPC-small & 2 & 2 & 256 & 12 & LS 100 hrs \\
        APC & 3 & 3 & 512 & 3 & LS 360 hrs \\
        \bottomrule
    \end{tabular}}
    \label{tab: model}
    \vspace{-3mm}
\end{table}

For both CPC models, the negative samples are drawn from the same speaker but not necessarily from the same utterance as the frame to be predicted.
It is possible that a CPC model that draws negative samples without speaker restrictions would be qualitatively different in terms of speaker and phone encoding, but we limit the current study to the within-speaker sampling case since it gives better phone classification results 
\cite{nguyen_zero_2020} and is thus the more common setup.

\subsection{Dataset}
We use data from LibriSpeech \cite{panayotov_librispeech_2015}, a corpus of English read speech. 
We perform our initial analyses on the \devclean\ portion, which contains 8 minutes of speech for each of 19 male and 21 female speakers.\footnote{The documentation claims that \devclean\ contains 20 males and 20 females, but through our analysis, we found that one of the ``male'' speakers (ID: 7976, name: JenniferRutters) appears to be female.}
For testing generalization, we extract
a speaker subspace from the \trainclean\ portion, which contains 25 minutes of speech for each of 126 male and 125 female speakers.
We then evaluate on \devclean\ and \testclean\ by collapsing
the speaker subspace learned from
\trainclean. 
The \testclean\ portion consists of 8-minute speech per speaker for 20 male and 20 female speakers.
The speakers occurred in the three sets are mutually disjoint. 

The phone labels required for our anlayses were obtained by performing forced alignment with an acoustic model created according to the official Kaldi recipe for LibriSpeech data\footnote{\url{https://github.com/kaldi-asr/kaldi/blob/master/egs/librispeech/s5/run.sh}}.
We ignore frames aligned to ``silence'' and
``spoken noise'' labels in our analysis, leaving 39 phone categories (i.e. $|P|=39$).

\subsection{Evaluation}
To evaluate how much speaker information is removed and its effect on the phonetic information in the representations, we perform two types of tests. 
First, we train speaker and phone {\bf probing classifiers}. 
We use linear classifiers, which are trained to predict either speaker or phone labels based on a single representation frame. 
We train each classifier on a random half of each speaker's utterances, using the other half for testing. We also use a machine ABX {\bf phone discrimination} test \cite{Schatz2013EvaluatingSF}. 
This test asks whether triphone $x$ is more similar to triphone $a$ than $b$, where $a$ and $x$ are tokens of the same type (\textit{e.g. `aba'}) and $b$ is of a different type (\textit{`apa'}).
The final ABX error rate is computed by aggregating the error rate with ($a$, $b$, $x$) iterating over all tokens combinations and triphone contrasts.
In \textit{within-speaker} ABX test, each triplet are tokens produced the same speaker.
In the \textit{across-speaker} setting, $a$ and $b$ are drawn from the same speaker and $x$ from a different speaker.
We follow the Zero Speech challenge splits \cite{nguyen_zero_2020} for within- and across-speaker ABX tests.
\subsection{Baselines for speaker normalization experiments} 
Previous work \cite{niekerk21_interspeech} has shown that utterance-level standardization (centering plus rescaling) effectively removes speaker information from CPC representations, improving performance on ABX tests and other tasks benchmarked in the Zero Speech 2021 challenge. 
We therefore use utterance-level standardization as a baseline, but also report centering alone, 
which we found to work better.
For analyses where data from all speakers is available in advance, we apply these approaches at the speaker level rather than the utterance level for a fair comparison to our methods.

\section{Analysis of subspaces}
\label{sec: analysis}

\begin{figure}
    \centering
    \includegraphics[width=0.9\linewidth]{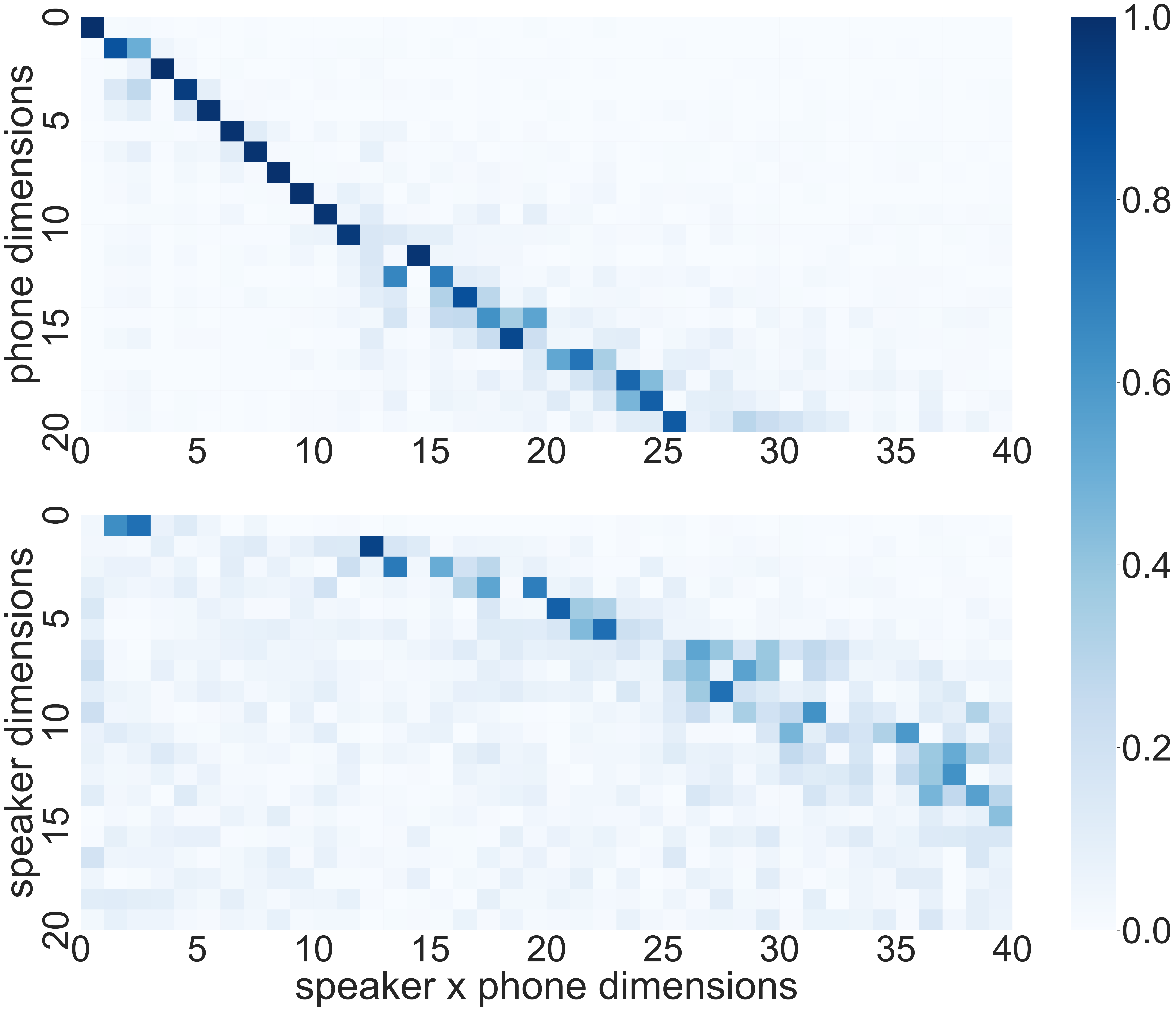}
        \caption{Similarity (absolute value of dot product) between the principal directions of
        $M_\text{phn}$ \& $M_{\text{joint}}$ (top)
        and  $M_\text{spk}$ \& $M_{\text{joint}}$ (bottom), computed from \texttt{dev-clean} using representations from CPC-big. 
        The patterns are similar for APC or CPC-small.} 
    \label{fig: similarity}
    \vspace{-3em}
\end{figure}

\begin{figure*}[b]
\centering
   \includegraphics[width=.9\textwidth]{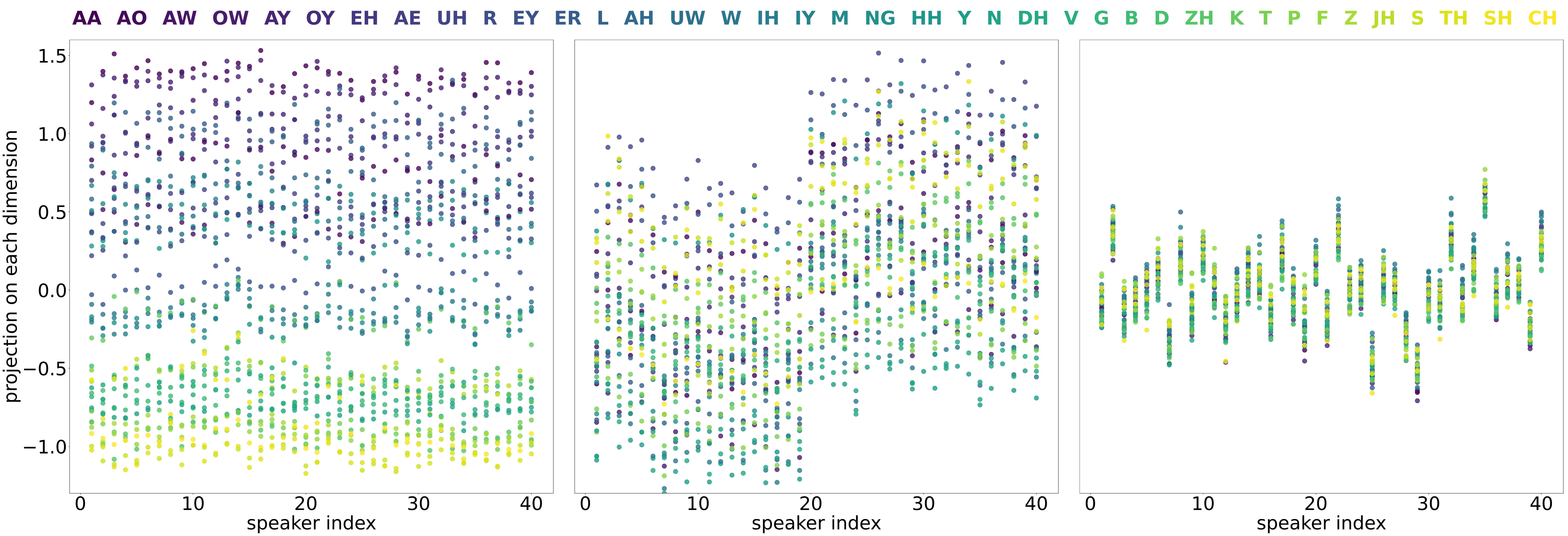}
\caption{Projection of $M_{\text{joint}}$ (extracted from \texttt{dev-clean} using CPC-big) onto its principal dimension 0 (left), 1 (middle), and 12 (right). 
Each dot represents a row in $M_{\text{joint}}$, with each speaker plotted in a single column and phone identities colour-coded.
The first 19 columns represent male speakers and the rest females.
The legend on top is ordered by each phone's average projection on dimension 0.
}
    \label{fig: projection}
\end{figure*}

To examine the relationship between speaker and phone
subspaces, we compute the 
similarity
between the top principal
directions of $M_\text{spk}, M_\text{phn},$ and $M_{\text{joint}}$, as extracted from \texttt{dev-clean}.
Since we only care about measuring orthogonality, we use the absolute value of the dot product as our similarity measure; this ranges from 0 to 1.

When comparing the principal directions of $M_\text{phn}$ and $M_\text{spk}$, we found very low similarity: e.g., for the CPC-big model, amongst the top 20 speaker directions, their similarity with the most aligned phone direction is on average only $0.13$ (variance: $0.002$, maximum: $0.26$),
indicating that the speaker and phone subspaces are nearly orthogonal.

We then compared the principal
directions of $M_\text{phn}$ and $M_\text{spk}$ to those of $M_{\text{joint}}$, as shown in Figure \ref{fig: similarity}. 
We see that, of the top 13 directions of $M_{\text{joint}}$, only two (directions 1 and 2) are similar to principal directions of both $M_\text{phn}$ and $M_\text{spk}$. Directions 0 and 3-11 of $M_{\text{joint}}$ align only to phone directions, while direction 12 aligns only to a speaker direction.
Moreover, while directions 1-2 are somewhat aligned with both speaker direction 0 and phone direction 1, this does not mean that speaker direction 1 is aligned with phone direction 0---in fact, their cosine similarity is only $0.169$.
Taken together, these observations further support the orthogonality of speaker and phone directions.

The preceding analysis
also informs us of the relative variance of phone and speaker encoding.
While $M_{\text{joint}}$ consists of combinations of 40 speakers and 39 phones, \textit{i.e.} roughly the same number of speaker and phone categories, 
most of the top principal
directions are used to encode phones.
That is, 
the SSL representations encode much less inter-speaker variance and use more
directions to discriminate phones than speakers.
This is not a surprise since the training objective of both models entail learning to discriminate between phones and not speakers.
What remains to be explained is how the models come to encode speaker information despite only being trained to predict or discriminate phones within each speaker.

To better understand what is encoded in some of the top principal
directions, we
visualize the projection of $M_{\text{joint}}$ onto its dimensions 0, 1, and 12 (Figure \ref{fig: projection}).
In the projection plot for dimension 0, we can see vowels lying on the positive end and fricatives and affricates lying on the negative end, 
indicating that this dimension discriminates phonetic categories along a sonority gradient while maintaining no information about speaker differences. 
Dimension 1, as well as dimension 2 which is not shown here, differentiate between genders while the consistent ordering of the coloured dots across all columns implies that they also capture some phonetic information.
In contrast, dimension 12 contains relatively greater inter-speaker variance, albeit still much smaller than variance between phones captured in dimension 0 and 1.
The projection on these dimensions are consistent with our observation from the similarity analysis.

\section{Application to speaker normalization}

\begin{table}
    \centering
    \caption{Probing (speaker \& phone) and ABX (within- \& across-speaker) error rates on \devclean.  
}
    \resizebox{\columnwidth}{!}{
    \begin{tabular}{llcccc}
    \toprule
    \multirow{2}{*}{Model}
    & \multirow{2}{*}{Error rate (\%)} & \multirow{2}{*}{Original} & \multicolumn{2}{c}{Speaker-level} & Speaker space  \\
    & & & centered & +rescaled & collapsed \\
    \midrule
\multirow{4}{*}{CPC-big}  & Speaker  & 0.45 &   76.07  & +6.92 & 82.30 \\
      &  Phone  &  25.18 & 25.19 & -0.73 & 25.37 \\
      &  ABX Within  & 3.38  &  3.39  & +0.02 &  \textbf{3.24}   \\
      &  ABX Across  & 4.11  & 3.97  & +0.07   & \textbf{3.77}  \\
    \midrule
     \multirow{4}{*}{CPC-small}  & Speaker  &  10.21  &    67.08  &  +6.59  & 84.69  \\
      &  Phone &  36.47 & 36.11  &  -0.72  & 37.20  \\
      &  ABX Within & 6.22  & 6.14 & -0.13 & \textbf{5.22}  \\
      &  ABX Across & 8.10 & 7.38 & +0.05 &  \textbf{6.73} \\
    \midrule
     \multirow{4}{*}{APC}  & Speaker  & 15.47 &  83.02 & +3.39  &  85.47 \\
      &  Phone & 36.29  & 36.15  & +0.00  &  36.75 \\
      &  ABX Within & 6.77 & 6.51 & +0.05 &  \textbf{6.45} \\
      &  ABX Across & 9.83 & \textbf{9.09} & +0.05  &  9.24  \\
    \bottomrule
    \end{tabular}
    }
    \label{tab: main}
    \vspace{-2mm}
\end{table}

\begin{figure}
    \centering
    \includegraphics[width=\linewidth]{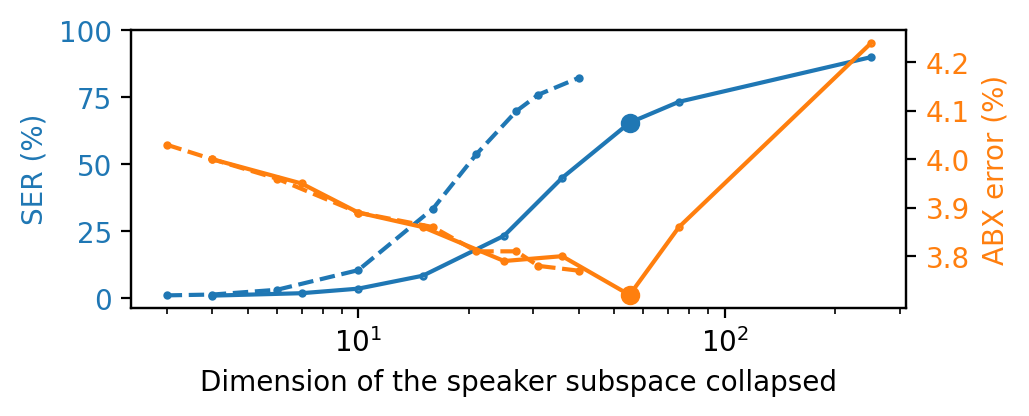}
    \caption{Speaker classification (blue) and across-speaker ABX (orange) on \devclean\ after collapsing speaker subspaces learned from either \trainclean\ (solid lines) or \devclean\ (dashed lines). The large dots are where the number of dimensions covers 95\% of the variance.  \label{fig: explained variance}}
    \vspace{-1.5em}
\end{figure}

\mypara{Collapsing speaker subspace of seen speakers}
We first explore speaker normalization for a set of known speakers, by learning the speaker subspace from \devclean, and collapsing the speaker directions on the same dataset. Since the number of speakers is small (40) relative to the number of dimensions (256 or 512), we collapse all 40 of the speaker directions. (See below for results with fewer directions collapsed.)
Table \ref{tab: main} reports results in comparison to speaker-level centering and standardization,
because here all speakers are known in advance. 
Utterance-level centering and standardization show similar results but are slightly less effective in removing speaker information.

\begin{table}
    \centering
    \caption{Results of speaker normalization on \devclean\ (top) and \testclean\ (bottom) by collapsing speaker subspaces learnt from \trainclean. 
    }
    \resizebox{\columnwidth}{!}{
    \begin{tabular}{cllcccc}
    \toprule
    & Model & Error (\%) & Original & Utt.\ centered & Collapsed
    \\\midrule
    \parbox[c]{0.3cm}{\multirow{10}{*}{\rotatebox{90}{\devclean}}}
    & \multirow{3}{*}{CPC-big}  
         & Speaker  &  0.45 & 69.04  & 65.52 \\
         & &  ABX Within  & 3.38 & 3.44 & \textbf{3.14}  \\
         & &  ABX Across  & 4.11 & 4.03 & \textbf{3.80}  \\\cmidrule{2-6}
    & \multirow{3}{*}{CPC-small}  
         & Speaker &  10.21 & 68.33 & 65.08 \\
         & &  ABX Within & 6.22 & 5.86 & \textbf{5.32}  \\
         & &  ABX Across &  8.10 & 7.26 & \textbf{6.91}\\\cmidrule{2-6}
    & \multirow{3}{*}{APC}  
          & Speaker & 15.47 & 76.37 &  69.48  \\
         & &  ABX Within  & 6.77 & \textbf{6.42} &  6.45 \\
         & &  ABX Across & 9.83 & \textbf{9.01} & 9.24   \\\midrule
    \parbox[c]{0.3cm}{\multirow{7}{*}{\rotatebox{90}{\testclean}}}
    & \multirow{2}{*}{CPC-big}  
         &  ABX Within  & 3.29 & 3.27 & \textbf{3.10} \\
         & &  ABX Across  & 4.22 & 4.11 & \textbf{4.01}\\\cmidrule{2-6}
    & \multirow{2}{*}{CPC-small}  
         &  ABX Within & 5.86 & 5.54 &  \textbf{4.85}\\
         & &  ABX Across & 7.48 & 6.91 & \textbf{6.37}\\\cmidrule{2-6}
    & \multirow{2}{*}{APC}  
         &  ABX Within  & 6.62 & \textbf{6.13} & 6.18 \\
         & &  ABX Across &   9.47 & \textbf{8.72}  & 8.98
         \\\bottomrule
    \end{tabular}
    }
    \label{tab: generalize}
    \vspace{-3mm}
\end{table}

Our method shows the best ABX results in nearly all cases, as well as close to 100\% speaker error rates and little change in phone error rates, indicating that this method removes nearly all the speaker information while improving the accessibility of phone information for unsupervised tasks. We also see from the baselines that removing speaker information need not {\em always} improve phone discrimination: in particular, applying rescaling on top of centering increases both the speaker error rate {\em and} the phone classification and ABX error rates.

\mypara{Collapsing speaker subspace of unseen speakers}
We first explore how results vary depending on the number of dimensions (and \% of variance) that are removed. 
Figure \ref{fig: explained variance} illustrates the results on CPC-big, comparing the case where the speaker subspace is learned and applied to the same speakers and the case where we apply it to new speakers.
When we have relatively few speakers as in \devclean, and we collapse the speaker subspace of the same speakers, including all of the speaker dimensions is helpful.
However, we see that when generalizing from the large number of speakers in \trainclean, many of the lower principal components appear to be overfitting: after about 50 dimensions, speaker error rate on the dev set increases only slowly while ABX error also begins to rise. 

Finally, after learning the speaker directions from the training set, we choose the number of directions to collapse based on the best across-speaker ABX scores on the development set, and present results for both the development and test sets in Table~\ref{tab: generalize}.
For CPC-big, CPC-small, and APC, we collapse speaker subspaces that explain 98\%, 95\%, and 95\% of training variance (57, 36, and 30 dimensions, respectively).
We compare to utterance-level centering, the baseline that gives the best results for the ABX task (of those applicable to previously unseen speakers). 

Both methods remove a similar amount of speaker information, but our method is better at the ultimate goal of improving phone discrimination in nearly all cases (the exception being the APC model on the test set). Moreover, it can be applied in a fully streaming setting since the speaker subspaces to collapse are computed in advance.

\section{Conclusions}

In analyses of SSL speech representations based on predictive coding models, we showed that speaker information and phonetic information are encoded in orthogonal dimensions of the representation space, indicating that these models are implicitly disentangling the two sources of information.
This insight led to a simple but effective speaker normalization technique that requires only speaker-labeled training data, and leads to improved phone discrimination on the test set.
Important avenues for future work include investigating the extent to which the speaker dimensions generalize to out of domain data (e.g., other languages or genres of speech), and whether the orthogonality we found extends to other SSL models based on different principles (e.g., masked prediction \cite{Hsu2021HuBERTSS, Baevski2020wav2vec2A}). 
\bibliographystyle{IEEEtran}
\bibliography{main}

\begin{thebibliography}{10}
\providecommand{\url}[1]{#1}
\csname url@samestyle\endcsname
\providecommand{\newblock}{\relax}
\providecommand{\bibinfo}[2]{#2}
\providecommand{\BIBentrySTDinterwordspacing}{\spaceskip=0pt\relax}
\providecommand{\BIBentryALTinterwordstretchfactor}{4}
\providecommand{\BIBentryALTinterwordspacing}{\spaceskip=\fontdimen2\font plus
\BIBentryALTinterwordstretchfactor\fontdimen3\font minus
  \fontdimen4\font\relax}
\providecommand{\BIBforeignlanguage}[2]{{%
\expandafter\ifx\csname l@#1\endcsname\relax
\typeout{** WARNING: IEEEtran.bst: No hyphenation pattern has been}%
\typeout{** loaded for the language `#1'. Using the pattern for}%
\typeout{** the default language instead.}%
\else
\language=\csname l@#1\endcsname
\fi
#2}}
\providecommand{\BIBdecl}{\relax}
\BIBdecl

\bibitem{Zhang2023GoogleUS}
Y.~Zhang, W.~Han, J.~Qin, Y.~Wang, A.~Bapna, Z.~Chen, N.~Chen, B.~Li,
  V.~Axelrod, G.~Wang, Z.~Meng, K.~Hu, A.~Rosenberg, R.~Prabhavalkar, D.~S.
  Park, P.~Haghani, J.~Riesa, G.~Perng, H.~Soltau, T.~Strohman, B.~Ramabhadran,
  T.~N. Sainath, P.~J. Moreno, C.-C. Chiu, J.~Schalkwyk, F.~Beaufays, and
  Y.~Wu, ``Google {USM}: {Scaling} {Automatic} {Speech} {Recognition} {Beyond}
  100 {Languages},'' \emph{ArXiv}, vol. abs/2303.01037, 2023.

\bibitem{polyak21_interspeech}
A.~Polyak, Y.~Adi, J.~Copet, E.~Kharitonov, K.~Lakhotia, W.-N. Hsu, A.~Mohamed,
  and E.~Dupoux, ``{Speech Resynthesis from Discrete Disentangled
  Self-Supervised Representations},'' in \emph{Proc. Interspeech 2021}, 2021,
  pp. 3615--3619.

\bibitem{blandon2020analysis}
M.~A.~C. Bland{\'o}n and O.~R{\"a}s{\"a}nen, ``Analysis of predictive coding
  models for phonemic representation learning in small datasets,'' in
  \emph{ICML 2020 Workshop on Self-supervision in Audio and Speech}, 2020.

\bibitem{lavechin2022statistical}
M.~Lavechin, M.~De~Seyssel, H.~Titeux, H.~Bredin, G.~Wisniewski, A.~Cristia,
  and E.~Dupoux, ``Can statistical learning bootstrap early language
  acquisition? {A} modeling investigation,'' 2022.

\bibitem{wells22_interspeech}
D.~Wells, H.~Tang, and K.~Richmond, ``{Phonetic Analysis of Self-supervised
  Representations of English Speech},'' in \emph{Proc. Interspeech 2022}, 2022,
  pp. 3583--3587.

\bibitem{sanabria_analyzing_2022}
R.~Sanabria, H.~Tang, and S.~Goldwater, ``Analyzing {Acoustic} {Word}
  {Embeddings} from {Pre}-trained {Self}-supervised {Speech} {Models},'' in
  \emph{2022 {IEEE} {International} {Conference} on {Acoustics}, {Speech} and
  {Signal} {Processing} ({ICASSP})}.\hskip 1em plus 0.5em minus 0.4em\relax
  IEEE, 2022.

\bibitem{deseyssel22_interspeech}
M.~{de Seyssel}, M.~Lavechin, Y.~Adi, E.~Dupoux, and G.~Wisniewski, ``{Probing
  phoneme, language and speaker information in unsupervised speech
  representations},'' in \emph{Proc. Interspeech 2022}, 2022, pp. 1402--1406.

\bibitem{niekerk21_interspeech}
B.~van Niekerk, L.~Nortje, M.~Baas, and H.~Kamper, ``{Analyzing Speaker
  Information in Self-Supervised Models to Improve Zero-Resource Speech
  Processing},'' in \emph{Proc. Interspeech 2021}, 2021, pp. 1554--1558.

\bibitem{pasad}
A.~Pasad, J.-C. Chou, and K.~Livescu, ``Layer-wise analysis of a
  self-supervised speech representation model,'' in \emph{2021 IEEE Automatic
  Speech Recognition and Understanding Workshop (ASRU)}, 2021, pp. 914--921.

\bibitem{chung19_interspeech}
Y.-A. Chung, W.-N. Hsu, H.~Tang, and J.~Glass, ``{An Unsupervised
  Autoregressive Model for Speech Representation Learning},'' in \emph{Proc.
  Interspeech 2019}, 2019, pp. 146--150.

\bibitem{oord_representation_2019}
A.~v.~d. Oord, Y.~Li, and O.~Vinyals, ``Representation {Learning} with
  {Contrastive} {Predictive} {Coding},'' \emph{arXiv:1807.03748 [cs, stat]},
  Jan. 2019, arXiv: 1807.03748.

\bibitem{Hsu2021HuBERTSS}
W.-N. Hsu, B.~Bolte, Y.-H.~H. Tsai, K.~Lakhotia, R.~Salakhutdinov, and
  A.~Mohamed, ``Hubert: Self-supervised speech representation learning by
  masked prediction of hidden units,'' \emph{IEEE/ACM Transactions on Audio,
  Speech, and Language Processing}, vol.~29, pp. 3451--3460, 2021.

\bibitem{Baevski2020wav2vec2A}
A.~Baevski, H.~Zhou, A.~rahman Mohamed, and M.~Auli, ``wav2vec 2.0: A framework
  for self-supervised learning of speech representations,'' \emph{ArXiv}, vol.
  abs/2006.11477, 2020.

\bibitem{elias1955predictive}
P.~Elias, ``Predictive coding--{I},'' \emph{IRE transactions on information
  theory}, vol.~1, no.~1, pp. 16--24, 1955.

\bibitem{yang2022autoregressive}
G.-P. Yang, S.-L. Yeh, Y.-A. Chung, J.~Glass, and H.~Tang, ``Autoregressive
  predictive coding: A comprehensive study,'' \emph{IEEE Journal of Selected
  Topics of Signal Processing}, 2022.

\bibitem{huang_predictive_2011}
Y.~Huang and R.~P.~N. Rao, ``\BIBforeignlanguage{en}{Predictive coding},''
  \emph{\BIBforeignlanguage{en}{WIREs Cognitive Science}}, vol.~2, no.~5, pp.
  580--593, 2011.

\bibitem{Rao1999PredictiveCI}
R.~P.~N. Rao and D.~H. Ballard, ``Predictive coding in the visual cortex: a
  functional interpretation of some extra-classical receptive-field effects.''
  \emph{Nature Neuroscience}, vol.~2, pp. 79--87, 1999.

\bibitem{Friston2005ATO}
K.~J. Friston, ``A theory of cortical responses,'' \emph{Philosophical
  Transactions of the Royal Society B: Biological Sciences}, vol. 360, pp. 815
  -- 836, 2005.

\bibitem{nguyen_zero_2020}
T.~A. Nguyen, M.~de~Seyssel, P.~Rozé, M.~Rivière, E.~Kharitonov, A.~Baevski,
  E.~Dunbar, and E.~Dupoux, ``The {Zero} {Resource} {Speech} {Benchmark} 2021:
  {Metrics} and baselines for unsupervised spoken language modeling,''
  \emph{arXiv:2011.11588 [cs, eess]}, Dec. 2020, arXiv: 2011.11588.

\bibitem{panayotov_librispeech_2015}
V.~Panayotov, G.~Chen, D.~Povey, and S.~Khudanpur,
  ``\BIBforeignlanguage{en}{Librispeech: {An} {ASR} corpus based on public
  domain audio books},'' in \emph{\BIBforeignlanguage{en}{2015 {IEEE}
  {International} {Conference} on {Acoustics}, {Speech} and {Signal}
  {Processing} ({ICASSP})}}.\hskip 1em plus 0.5em minus 0.4em\relax South
  Brisbane, Queensland, Australia: IEEE, Apr. 2015, pp. 5206--5210.

\bibitem{Schatz2013EvaluatingSF}
T.~Schatz, V.~Peddinti, F.~R. Bach, A.~Jansen, H.~Hermansky, and E.~Dupoux,
  ``Evaluating speech features with the minimal-pair abx task: analysis of the
  classical mfc/plp pipeline,'' in \emph{Interspeech}, 2013.

\end{thebibliography}

\end{document}